\documentclass{article}

\usepackage{microtype}
\usepackage{graphicx}
\usepackage{subcaption}
\usepackage{booktabs} % for professional tables

\usepackage{hyperref}

\usepackage{amsmath,amsfonts,bm}

\def\eqref#1{equation~\ref{#1}}
\def\1{\bm{1}}

\newcommand{\test}{\mathcal{D_{\mathrm{test}}}}

\def\vtheta{{\bm{\theta}}}

\def\vx{{\bm{x}}}

\DeclareMathAlphabet{\mathsfit}{\encodingdefault}{\sfdefault}{m}{sl}
\SetMathAlphabet{\mathsfit}{bold}{\encodingdefault}{\sfdefault}{bx}{n}

\newcommand{\Ls}{\mathcal{L}}

\DeclareMathOperator*{\argmax}{arg\,max}

\DeclareMathOperator{\sign}{sign}

\usepackage{enumitem}

\usepackage[arxiv]{icml2024}

\usepackage{amsmath}
\usepackage{amssymb}
\usepackage{mathtools}
\usepackage{amsthm}

\usepackage[capitalize,noabbrev]{cleveref}

\theoremstyle{plain}

\theoremstyle{definition}

\theoremstyle{remark}

\usepackage[textsize=tiny]{todonotes}

\icmltitlerunning{PUMA: margin-based data pruning}

\begin{document}

\twocolumn[
\icmltitle{PUMA: margin-based data pruning \\ for better accuracy-robustness trade-offs}



\icmlsetsymbol{equal}{*}

\begin{icmlauthorlist}
\icmlauthor{Javier Maroto}{yyy}
\icmlauthor{Pascal Frossard}{yyy}
\end{icmlauthorlist}

\icmlaffiliation{yyy}{EPFL, Switzerland}

\icmlcorrespondingauthor{Javier Maroto}{javier.marotomorales@epfl.ch}

\icmlkeywords{Adversarial robustness, data pruning}

\vskip 0.3in
]



\printAffiliationsAndNotice{}  % leave blank if no need to mention equal contribution

\begin{abstract}
Deep learning has been able to outperform humans in terms of classification accuracy in many tasks. However, to achieve robustness to adversarial perturbations, the best methodologies require to perform adversarial training on a much larger training set that has been typically augmented using generative models (e.g., diffusion models). Our main objective in this work, is to reduce these data requirements while achieving the same or better accuracy-robustness trade-offs. We focus on data pruning, where some training samples are removed based on the distance to the model classification boundary (i.e., margin). We find that the existing approaches that prune samples with low margin fails to increase robustness when we add a lot of synthetic data, and explain this situation with a perceptron learning task. Moreover, we find that pruning high margin samples for better accuracy increases the harmful impact of mislabeled perturbed data in adversarial training, hurting both robustness and accuracy. We thus propose PUMA, a new data pruning strategy that computes the margin using DeepFool, and prunes the training samples of highest margin without hurting performance by jointly adjusting the training attack norm on the samples of lowest margin. We show that PUMA can be used on top of the current state-of-the-art methodology in robustness, and it is able to significantly improve the model performance unlike the existing data pruning strategies. Not only PUMA achieves similar robustness with less data, but it also significantly increases the model accuracy, improving the performance trade-off.\looseness=-1
% TODO: check more benefits maybe
\end{abstract}

\section{Introduction}
\label{sec:intro}

The ability of deep learning models to generalize has been attributed not only to the large computation capabilities, but also to the increasing availability of data~\cite{krizhevsky2012imagenet}. Recently, obtaining models that do not only generalize but are also adversarially robust has been a challenge~\cite{xing2021adversarially}. Adversarial training is currently the best approach to obtain such robust models but it suffers from robust overfitting~\cite{rice2020overfitting}, a pronounced decrease in accuracy and robustness in the late stages of training~\cite{rice2020overfitting}. Recently, it has been observed that this problem can be avoided by training the model with even larger quantities of data, which can be obtained from generative models trained on the original data~\cite{rebuffi2021fixing,xing2022artificially,wang2023better}%\footnote{For a fair comparison with methods that do not use extra data, they train their diffusion models with the original training set.}. 
Particularly, for the CIFAR10 dataset, adversarial training continues improving even after generating 50 million examples with complex diffusion models~\cite{wang2023better} (i.e., 1000 times the original training set size). This increase in data quantity increases significantly the training time, much beyond the inherent cost of classical adversarial training, making the learning of robust models very computationally demanding.

\begin{table}[t]
    \centering
    \begin{tabular}{c|cc}
        \toprule
        & Accuracy & Robustness \\
        \hline
        SOTA method & $87.87_{\pm .10}$ & $\textbf{58.57}_{\pm .08}$ \\
        SOTA method using PUMA & $\textbf{91.37}_{\pm .21}$ & $\textbf{58.53}_{\pm .36}$ \\
    \end{tabular}
    \caption{Improvement in performance of the SOTA ResNet-18 in the robustness literature~\cite{wang2023better} when we apply our proposed pruning strategy, PUMA. We use the CIFAR10 dataset with 1M data samples generated using the EDM diffusion model~\cite{karras2022elucidating}.}
    \label{tab:summary}
\end{table}

Even before data scalability was a major issue, some works advocated to prune the training data to reduce computational cost, reduce labelling cost and speed up convergence. This strategy follows the natural intuition that there are some samples that could be more informative and others that may be redundant. Some works have shown that to maximize the model accuracy, it is preferable to only keep the samples that are most probably misclassified during training~\cite{scheffer2001active,balcan2007margin} (i.e., prune easy, or PE). Other works have pointed that this strategy is very detrimental to the model robustness when used in combination with adversarial training, and advocate for the opposite, pruning the most probably misclassified samples instead (i.e., prune difficult, or PD). They found that robustness may be increased, but with some reduction in accuracy~\cite{dong2021data,liu2021impact}. Notably, the PD strategy requires measuring how often each training sample is correctly classified during training, which is not feasible when millions of generated data points are added to the training set. Both strategies seem to show that pruning is unable to improve the trade-off between accuracy and robustness~\cite{tsipras2018robustness}, even in tasks like image classification, where we know that humans are capable to solve that task robustly and accurately. 

In this work, our objective is to design an effective data pruning strategy that is able to improve the accuracy-robustness trade-off, particularly when we deal with large training sets augmented with synthetic samples, as the ones used by current state-of-the-art robust models. We find that, in this large data regime, the PD strategy does not only hurt accuracy, but no longer improves robustness. To explain it, we construct a toy experiment with perceptron learning as proposed in~\cite{sorscher2022beyond} where they study the effect of the PD and PE strategies as we increase the training data size. In their experiments, the authors of~\cite{sorscher2022beyond} showed that the PE strategy outperforms the PD strategy for large quantities of training samples and that, compared to no pruning, the error scaling with the training size is exponential instead of a power law. However, instead of standard training, we use adversarial training at different training attack strengths (i.e., $\varepsilon$ distance) in this work. Not only it shows that the PD strategy no longer works when the dataset is large enough, but it also shows that the PE strategy does not work well in adversarial training. As we increase the norm of the adversarial examples used to train the model, we find that the samples with margin lower than the adversarial strength are very detrimental to the performance of the model. The core of our proposal is to recover the fast scaling that the PE strategy exhibited in non-adversarial settings, but for adversarial training. In fact, we show that it is possible on the perceptron learning problem when we remove the low-margin and high-margin samples simultaneously.\looseness=-1

However, when considering practical scenarios with real image data, removing those samples is not as simple, since we no longer know the true margin of each sample. Instead, we propose to use DeepFool~\cite{moosavi2016deepfool} to compute a proxy of the model margin. It can be used to determine which samples could be harmful and which ones should be removed with the PE strategy. Unlike in our previous experiment, we found that removing the potentially harmful samples is not very effective in real settings, so instead we propose to use a per-sample $\varepsilon_i$ training attack norm which depends on the sample margin to jointly minimize the average perturbed sample margin and the number of mislabeled perturbed samples. While there are other works that adaptively vary this per-sample $\varepsilon_i$ distance using heuristics during training~\cite{balaji2019instance,cheng2020cat}, these approaches are not practical in the large-data regime where most data is used once. We finally propose PUMA (i.e., PrUning MArgin), a new methodology that removes the training samples with highest margin, and simultaneously reduces the adversarial strength of the samples with lowest margin during adversarial training. We show that unlike other pruning strategies, it surpasses state-of-the-art performance, increasing the model accuracy without hurting its robustness. We show extensive ablation studies of the different hyperparameters used in the implementation, and confirm that our methodology generalizes to different architectures and training sizes. In summary, our contributions are:\looseness=-1
\begin{itemize}[itemsep=1pt,topsep=1pt,partopsep=0pt]%[nosep][topsep=8pt,itemsep=4pt,partopsep=4pt, parsep=4pt]
    \item We give a simple theoretical model that shows why existing pruning strategies are ineffective in the large-data regime adversarial setting. Moreover, it can give insights of when other proposed adversarial methods that change the training data or the adversarial strength could be ineffective.
    \item We show that the DeepFool model margin metric is a better proxy metric than the classification difficulty used by other pruning strategies, and better suited for the large-data regime.
    \item We propose a new algorithm, PUMA, that prunes based on the margin. It is adapted to the large-data regime adversarial setting and is able to improve significantly the accuracy of state-of-the-art robust models
\end{itemize}

We find PUMA empirically shows that data is as important as the architecture or optimization procedure in the general goals of deep learning. We think that our insights are general, and the increase in performance due to pruning will increase as the proxy metric of the true margin improves. Moreover, we believe it can lead to better and more efficient data generation, in which generative models will be purposely biased to creating samples with lower margin which can be used effectively even in adversarial settings.\looseness=-1

%We believe that this work can lead to better data generation, in which a diffusion model can be used to generate only samples that are neither too difficult, which risks adding label noise, neither too easy, which results in lower accuracy and potentially lower robustness as well. Moreover, this result is very promising to many other applications, where the margin metric should be easily translatable to other error metrics that are specific to the problem in question.

\section{Pruning analysis using perceptron learning}
\label{sec:perceptron_learning}

In this Section, we will use the well-defined perceptron learning task to study the optimality of the PD and PE pruning strategies in the context of adversarial training. The purpose is to understand why the PD strategy performs worse in the large data regime, and use the insights to design a better pruning strategy.

Perceptron learning in the student-teacher setting~\cite{gardner1988space,seung1992statistical} is an application of statistical mechanics~\cite{engel2001statistical, advani2013statistical}, that has been used to analyze machine learning problems. In this setting, the teacher generates random i.i.d. inputs $\vx_i = \mathcal{N}(0, 1)^K$, and their corresponding labels $y_i = \text{sign}(\vtheta_t \cdot \vx_i)$, where $\vtheta_t$ are the teacher weights that the student tries to learn. When giving these inputs to the student model, its test error scales with the size of the training set as a power law of value -1. Interestingly, it is possible to greatly improve how well the performance scales with the quantity of data by using active learning~\cite{ducoffe2018adversarial} or data pruning~\cite{sorscher2022beyond}. In the latter case, show that pruning the samples with highest margin can reduce this power law limit to exponential scaling. This is promising to reduce the data and computational requirements in real tasks, which also exhibit power law scaling.

In this work, we will use perceptron learning to analyze the effect of pruning in the context of adversarial training, and try to achieve exponential scaling as well. Our settings are the following. We will feed the student with adversarial examples computed using the FGSM attack~\cite{goodfellow2014explaining}, enclosed inside the L$_{\infty}$ ball of size $\varepsilon$. That is, given the teacher generated inputs $\vx_i$, their corresponding adversarial samples would be of the form $\vx_i' = \vx_i + \varepsilon \sign(\nabla_\vx \Ls (\vtheta_s \cdot \vx_i))$, where $\vtheta_s$ are the weights of the student at that instance of training. We point out that these adversarial examples are only designed to fool the student model and may not be optimal to fool the teacher. For the teacher, we will use $\vtheta_t = (1, 0, 0, ...)$ without loss of generality, and set the number of features $K = 200$. We define $\alpha$ as the ratio between the number of training samples and $K$. We will consider the two prevalent pruning strategies: either remove the highest margin (PE strategy) or the lowest margin samples (PD strategy). Moreover, we will vary the quantity of data pruned with those strategies. For the sake of fairness, when comparing the performance of the different strategies we make sure their training sizes are similar after pruning. We will also compare with a baseline where no data is pruned.\looseness=-1

\begin{figure}[t]
    \begin{subfigure}{0.49\linewidth}
        \centering
        \includegraphics[width=\linewidth]{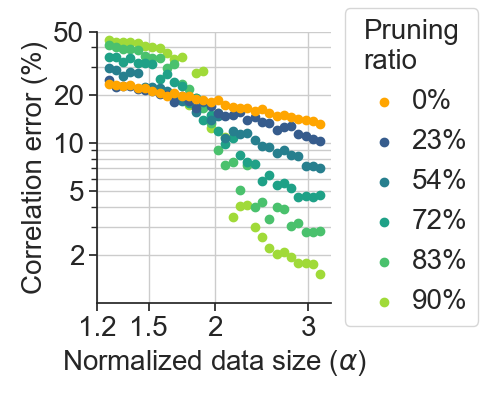}
        \caption{PE strategy}
        \label{fig:rs_pe_e0}
    \end{subfigure}
    \begin{subfigure}{0.49\linewidth}
        \centering
        \includegraphics[width=\linewidth]{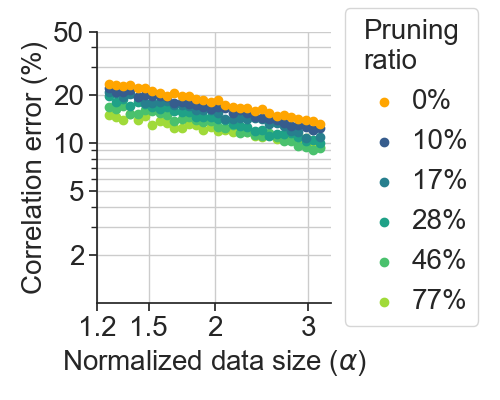}
        \caption{PD strategy}
        \label{fig:rs_pd_e0}
    \end{subfigure}
    \caption{Correlation error between the teacher and the student weights relative to $\alpha$ when standardly training the student. The orange line shows the performance when not pruning. For $\alpha < 2$, only the PD strategy improves performance. For $\alpha > 2$, as we increase the pruning ratio, the PE strategy outperforms significantly not pruning and has exponential scaling instead of power law.}
    \label{fig:rs_eps0}
\end{figure}

Setting $\varepsilon = 0$ replicates the results obtained in~\cite{sorscher2022beyond}, where they use standard training. It also sets a baseline to compare for the following experiments, in which we will increase $\varepsilon$. We show in Figure \ref{fig:rs_eps0} the error in the correlation between the teacher and student weights with respect to $\alpha$. In orange, we show the pruning law relationship of not pruning. While the PD strategy is a bit better than not pruning (Figure \ref{fig:rs_pd_e0}), it still follows a power law no matter how many samples are pruned. On the contrary, the PE strategy is only better for high enough $\alpha$. This is intuitive, because despite the low margin samples being the most informative, they can also induce overfitting when they are too few.\looseness=-1

\begin{figure}
    \begin{subfigure}{0.49\linewidth}
        \centering
        \includegraphics[width=\linewidth]{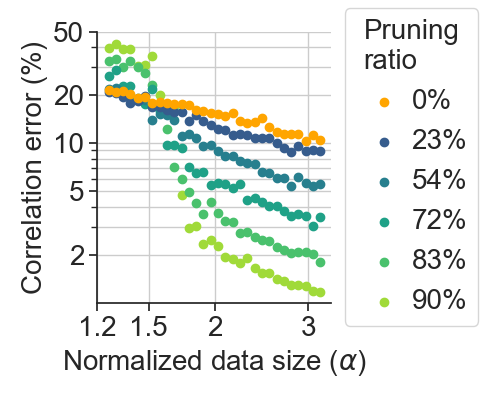}
        \caption{PE strategy}
    \end{subfigure}
    \begin{subfigure}{0.49\linewidth}
        \centering
        \includegraphics[width=\linewidth]{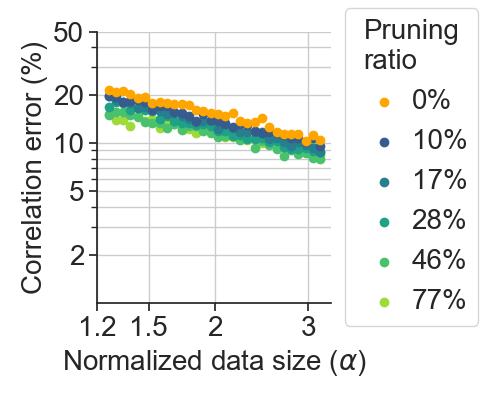}
        \caption{PD strategy}
    \end{subfigure}
    \caption{Correlation error between the teacher and the student weights relative to $\alpha$ when adversarially training the student with $\varepsilon = 0.001$.\looseness=-1}
    \label{fig:rs_eps0.001}
\end{figure}

In Figure \ref{fig:rs_eps0.001} we increased slightly $\varepsilon$ and find that adversarial training does in fact help the student performance, reducing the correlation error between teacher and student weights by 3-4\% lower correlation error for all $\alpha$'s. Intuitively, adversarial training consists in augmenting the training data, which in turns reduces overfitting and artificially increases the data size. In real data, we see this observation is consistent with the increased accuracy and robustness results obtained empirically when adversarial training with very small values of $\varepsilon$ on ImageNet~\cite{xie2020adversarial,bochkovskiy2020yolov4}

\begin{figure}
    \begin{subfigure}{0.49\linewidth}
        \centering
        \includegraphics[width=\linewidth]{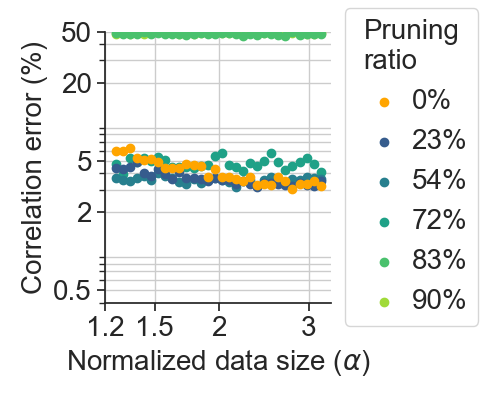}
        \caption{PE strategy}
    \end{subfigure}
    \begin{subfigure}{0.49\linewidth}
        \centering
        \includegraphics[width=\linewidth]{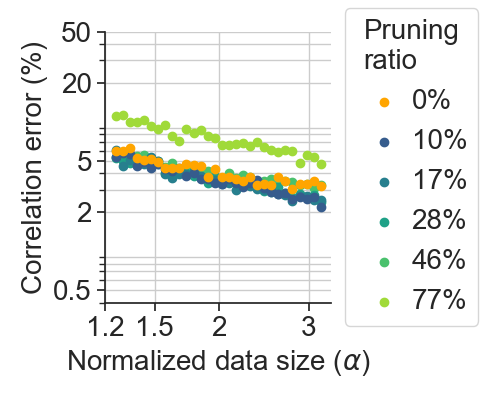}
        \caption{PD strategy}
    \end{subfigure}
    \caption{Correlation error between the teacher and student weights relative to $\alpha$ when adversarially training the student with $\varepsilon = 0.01$.\looseness=-1}
    \label{fig:rs_eps0.01}
\end{figure}

\begin{figure}
    \begin{subfigure}{0.49\linewidth}
        \centering
        \includegraphics[width=\linewidth]{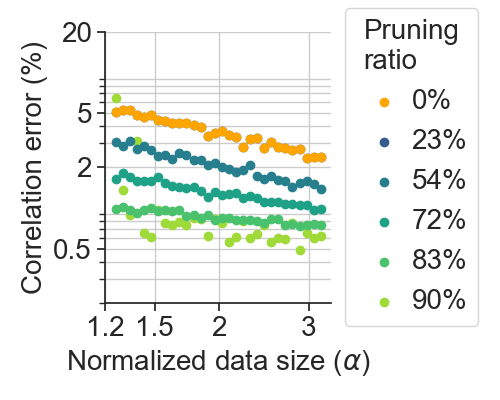}
        \caption{PE strategy}
    \end{subfigure}
    \begin{subfigure}{0.49\linewidth}
        \centering
        \includegraphics[width=\linewidth]{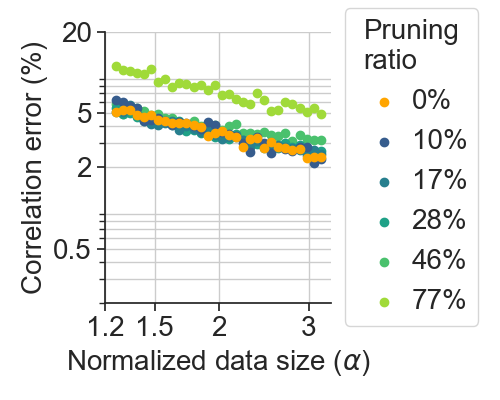}
        \caption{PD strategy}
    \end{subfigure}
    \caption{Correlation error between the teacher and the student weights relative to $\alpha$ when adversarially training the student with $\varepsilon = 0.01$, after filtering all the samples with margin lower than $\varepsilon$.}
    \label{fig:rs_eps0.01_f}
\end{figure}

\begin{figure*}[!t]
    \centering
    \includegraphics[width=0.7\linewidth]{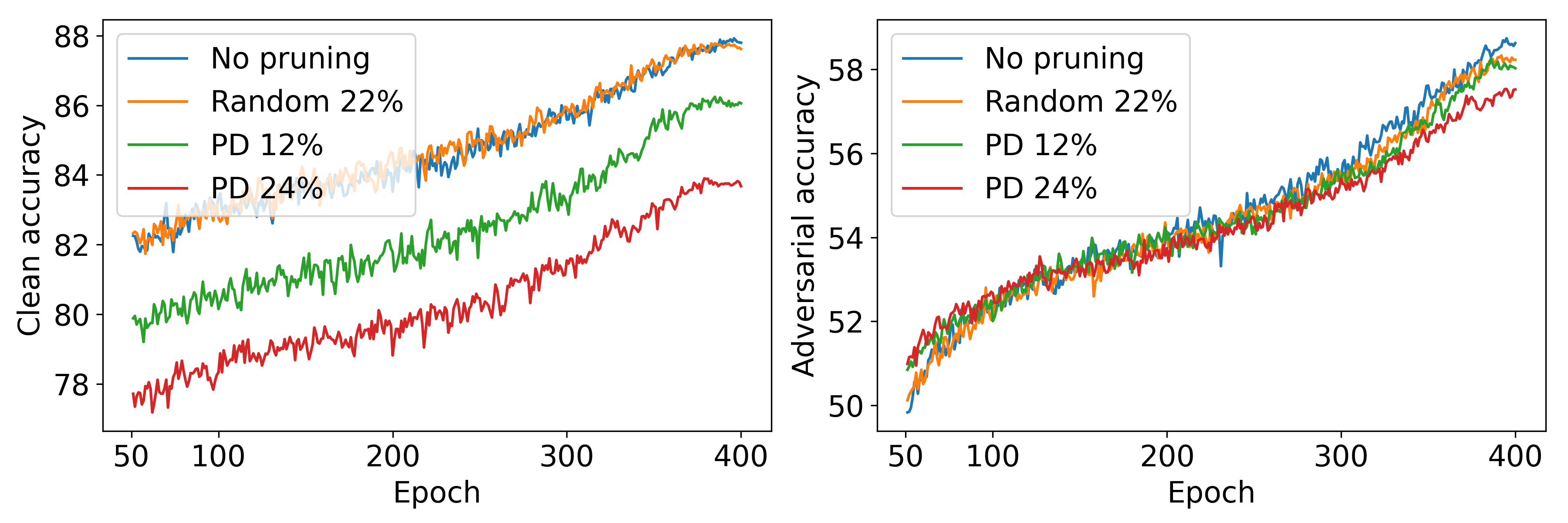}
    \caption{PD strategy for the ResNet-18 model adversarially trained on CIFAR10 with 1M EDM-generated samples. At the beginning (i.e. first 100-150 epochs), the model still has not seen enough new images and pruning improves robustness. But as the model trains with more new images, its robustness decreases compared to random pruning. Moreover, accuracy is significantly worse in all training stages.}
    \label{fig:prune_difficult}
\end{figure*}

\begin{figure*}[t]
    \centering
    \includegraphics[width=0.7\linewidth]{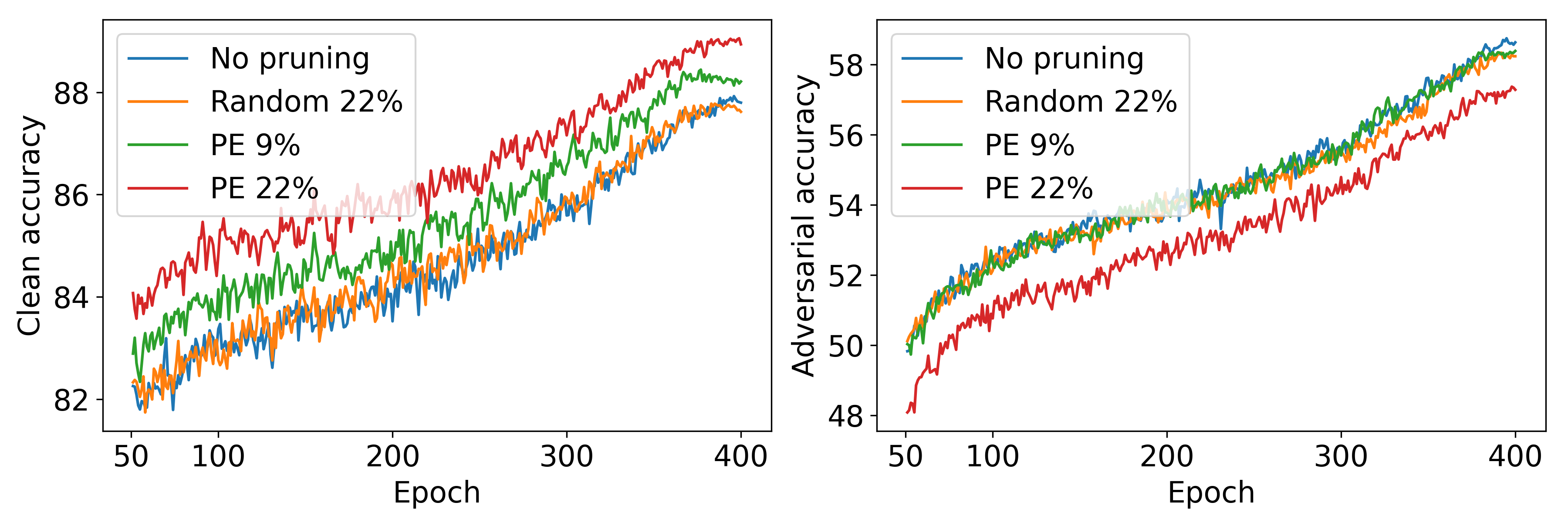}
    \caption{PE strategy for the ResNet-18 model adversarially trained on CIFAR10 with 1M EDM-generated examples. Consistent with previous results by other works, it has lower robustness compared to similarly sized random pruning. However, it outperforms the PD strategy at the end of the training. Moreover, its accuracy is significantly better in all training stages.}
    \label{fig:prune_easy}
\end{figure*}

However, to explain why the PE and PD strategies do not work with adversarial training on the large data regime we need to increase $\varepsilon$ further. Adversarial perturbations are large enough that they significantly decrease the margin of the data, which is reflected in a large performance improvement when not pruning for all $\alpha$'s\footnote{This trend is reversed at even higher $\varepsilon$ values due to the large proportion of mislabeled perturbed samples}. Moreover, if the sample had originally a very small margin, the adversarial attack can make it cross the true class decision boundary inducing label noise. That is, they are no longer correctly labelled. This is commonly referred as the invariance-sensitivity trade-off~\cite{tramer2020fundamental} in the adversarial literature.
On the one hand, this is the main reason why the PE strategy no longer follows exponential law. In Figure \ref{fig:rs_eps0.01} we show that, in fact, it starts underperforming compared to not pruning as we increase $\alpha$ and the pruning ratio. By pruning the samples with highest margin, the proportion of samples that are mislabelled increases, which heavily hinders the learning process of the student.
On the other hand, we show in Figure \ref{fig:rs_eps0.01} that the PD strategy still follows power law and is unaffected by the mislabelled samples. However, by removing the samples with low margin the overall margin increases. At high enough $\varepsilon$ and $\alpha$, it has a significant impact in performance, resulting in lower or similar performance compared to not pruning.\looseness=-1

Based on the previous results, to achieve exponential scaling, we need to solve the mislabelled samples issue in the PE strategy. A simple solution we have in this task is to also prune the samples with margin lower than $\varepsilon$. We show in Figure \ref{fig:rs_eps0.01_f} that this results in significantly better performance, following the same exponential scaling observed in Figures~\ref{fig:rs_eps0} and \ref{fig:rs_eps0.001}. Also, as expected, our solution has no effect in the PD strategy because it already prunes all low-margin samples (including the mislabelled ones).

\section{Pruning in image classification}

\subsection{General settings}

In the following Sections, we will use the insights obtained with perceptron learning, design an pruning algorithm that does not rely on the true margin of the data, and apply it to a more popular and complex task: image classification.\looseness=-1

We will use the same settings through the following experiments, unless otherwise stated. We use state-of-the-art methodology to adversarially train a model, as in~\cite{wang2023better}, and replicate their settings. That is, we use the CIFAR10 dataset and we add 1M data examples generated using the elucidating diffusion model (EDM). We train a ResNet-18 using TRADES and weight averaging. We train the network for 400 epochs where each epoch is comprised of 50K samples, 70\% selected randomly from the 1M generated examples, and 30\% from the original training set. The test robustness is reported using C\&W~\cite{carlini2017towards} attacks in all the following Figures and Tables. For the Tables, we will report the last epoch results.

For the pruning strategies, we consider the aforementioned prune difficult (PD) and prune easy (PE) strategies, which remove the most or the least difficult to classify samples, respectively. In the previous Section, we defined the most difficult examples as the examples with lowest margin, but other metrics can be used to order the examples by difficulty. Depending on the metric used, either a model has to be trained (e.g., classification difficulty) or a pretrained model has to be used (e.g., our proposed model margin metric). In both cases, we will use an adversarially trained model with the same architecture and the same training data as the one to be trained with the pruning strategy, unless otherwise stated. We define pruning ratio as the quantity of samples relative to the total that are removed from the training data by the pruning strategy.

\subsection{Proxies for the true margin}
% Explain why true margin cannot be used
% Current PD method is used for robustness, uses CD instead of true margin

While margin has proved to be a useful metric to determine which samples we want to prune in the perceptron learning task, computing the margin in computer vision is in the same manner is unfeasible because there is no real ground truth that we can use to measure the distance to the class boundary. For that reason, other works have proposed to use proxy metrics to prioritize samples~\cite{ducoffe2018adversarial,wang2021convergence}.

%The authors in~\cite{ducoffe2018adversarial} have successfully computed the margin to the model boundary with DeepFool~\cite{moosavi2016deepfool} and have used this metric for the problem of active learning, where they determine that the samples with lowest margin are the most important to be labelled first.

One such metric is the EL2N score~\cite{paul2021deep}, which is the average L$_2$ norm of the error vector of a small ensemble of models that has been trained for a very short time. The authors employed the PE strategy, and showed that, when standardly training on CIFAR10, they can remove up to 50\% of the lowest scoring training samples without loss of accuracy.
In contrast, the authors in~\cite{dong2021data,liu2021impact} showed that, for adversarial training in CIFAR10, it is better to use the PD strategy to maximize robustness. Instead of the EL2N score, they use the classification difficulty (CD) score to prioritize samples. Like~\cite{dong2021data}, we will use the following definition:
\begin{equation}
    \text{CD}_i = \dfrac{1}{T} \sum_{t=1}^T \argmax_k f_{tk}(\vx_i) \neq y_i
\end{equation}
where $\text{CD}_i$ is the classification difficulty of the sample $i$, and $f_{tk}$ is the value of the logit of the class $k$ computed by a model that has been trained adversarially for $T$ epochs.

\subsection{Performance of existing pruning strategies}
% Show that PD underperforms at the late stages of training (larger alpha)
% Show that PE is costly in terms of robustness, but increases accuracy
% Show that removing low CD samples with PE is not effective

In this work, we want to use pruning strategies effectively to reduce the large amount of data required and increase the accuracy-robustness trade-off of the state-of-the-art robust methodologies, which make use of diffusion models to augment the training data. %Moreover, we want to achieve a good accuracy-robustness trade-off.
Thus, our first step is to evaluate the performance of the existing PD and PE strategies when there are large quantities of data. On the one hand. based on the respective work claims, the PE strategy should increase accuracy at the cost of robustness, and the PD strategy should increase robustness at the cost of accuracy. On the other hand, the results we obtained with perceptron learning in Section~\ref{sec:perceptron_learning} suggest that the PD strategy should not longer be effective in the large data regime we are considering. Thus, in this Section, we will revisit and evaluate the performance of the prevalent pruning strategies in the context of adversarial training with large quantities of data generated with diffusion models.\looseness=-1

In Figure \ref{fig:prune_difficult}, we show that the insights obtained with the perceptron model translate for the CIFAR10 dataset with generated data. We compare the prevalent PD strategy on the bigger dataset with random or no pruning, and find that as the model trains with more images, the PD strategy effectiveness decreases. At the beginning, with bigger pruning ratios we can see a significant improvement in robustness compared to random pruning, confirming the results obtained by previous works on relatively small data set sizes. However, at the end of the training, when all the images have been processed, the larger the pruning ratio the more it underperforms in robustness compared with random pruning. Finally, we show that the PD strategy decreases the accuracy of the model, independently of the number of images processed, confirming that the difficult samples are the ones that help maximize the model accuracy.

In contrast, in Figure \ref{fig:prune_easy} we test the PE strategy. Compared to the PD strategy, we no longer see this inconsistency on the performance at different stages of training. Moreover, this strategy seems more attractive because of the increased accuracy compared to not pruning, even if it comes at the cost of some robustness. However, based on the insights obtained in the perceptron example, we believe this strategy can be further improved. After all, we have shown in the perceptron learning task that a much better strategy is to also remove the samples with margin lower than $\varepsilon$, improving the scaling performance from power law to exponential.

Before describing our proposed method, we show the results of directly translating our proposed strategy on the perceptron learning task in image classification. That is, using the PE strategy, but also removing the most difficult examples (i.e., CD = 1) which, problematically, represent 12\% of the entire data. Moreover, because we no longer have a ground truth, we have to make the assumption that these examples will tend to be closer to the true classification boundary. In practice, we find that by removing them, we are also removing a lot of samples that have small margins that do not cross the boundary when perturbed. This is reflected on Table \ref{tab:double_pruning}, where we can see that removing the most difficult samples results in much lower accuracy. While it is clearly not optimal, since it has similar performance to pruning random samples, it has better accuracy than the PD strategy with similar robustness. Thus, showing the potential of the obtained insights.\looseness=-1

\begin{table}[t]
    \centering
    \begin{tabular}{c|cc}
    \toprule
        & Accuracy & Robustness \\
        \hline
        No pruning & $87.87_{\pm .10}$ & $\textbf{58.57}_{\pm .08}$ \\
        \hline
        Random 22\% & $87.64_{\pm .04}$ & $58.27_{\pm .05}$ \\
        PD 12\% & 86.05$_{\pm .10}$ & $\textbf{58.40}_{\pm .19}$ \\
        PE 22\% & $\textbf{88.87}_{\pm .10}$ & $57.40_{\pm .17}$ \\
        PE 22\% + PD 12\% & $87.13_{\pm .10}$ & $\textbf{58.39}_{\pm .22}$ \\
    \end{tabular}
    \caption{Performance when using simple pruning strategies for the ResNet-18 model adversarially trained on CIFAR10 with 1M EDM-generated examples. We use CD to prioritize the pruning. This results in a pruning ratio of 22\% for the PE strategy by pruning the samples with CD $<$ 0.02, and 12\% for the PD strategy when pruning the samples with CD = 1. We show that a naive approach that combines PE and PD outperforms the PD strategy, but has similar performance than random pruning.}
    \label{tab:double_pruning}
\end{table}

\section{PUMA: pruning based on sample margin}
\label{sec:puma}

\subsection{Design choices}
% Per-sample epsilon: Motivate from previous results, (i.e. we want low margin and no mislabelling), our proposal of adjust the eps_i based on the margin
% DeepFool margin: Explain how and when the margin is computed, why we use a robustly pretrained model for computing this margin.

Motivated by the theoretical exponential scaling we have achieved in the large-data regime for adversarial training in the perceptron learning task, we propose PUMA, a new pruning strategy designed for, but not limited to, image classification. In PUMA, we first use a pretrained robust model to compute the model margin with DeepFool~\cite{moosavi2016deepfool}. We will use the term ``margin" to refer to this metric, since the true margin to the class boundary cannot be computed. Then, depending on the pruning ratio chosen, we remove the samples with the highest margin values from the training data. Finally, we perform adversarial training, but we vary the training attack norm, $\varepsilon$, depending on the sample margin value.

\paragraph{Model margin} The DeepFool algorithm has been design with the whole purpose of finding the closest L$_p$ adversarial examples for a particular model and set of data. Thus, it can compute the model margin for any training samples and any particular L$_p$ norm distance. DeepFool iteratively searches for the closest model boundary using gradient steps, followed by another set of iterative steps to ensure we are not overshooting the margin. For PUMA, we will always use adversarially robust models, because the margin values tend to be larger and it has a closer similarity to the CD metric, which uses an adversarially trained model as well. However, unlike the CD metric, it is not necessary to adversarially train a model from scratch, and a pretrained one (e.g. from RobustBench) or a finetuned one will suffice. Moreover, the model margin helps us identify when an adversarial example may cross the true class boundary, since we can compare the norm of the adversarial example directly with the metric. It is also more fine-grained that CD, in which many training samples may share the same CD value\footnote{For empirical results supporting our choice, see Appendix~\ref{sec:cd_margin_comparison}}.\looseness=-1 %This is especially useful when choosing an specific $\varepsilon$ value for each training sample, where using exclusively the CD metric limits greatly the performance (see Table~\ref{tab:margin_difficult}).

\paragraph{Adaptive $\varepsilon$} To improve the potential of the PE strategy, we saw that it is very important to avoid training with samples that are mislabeled. We propose to adjust the adversarial strength accordingly to avoid training with mislabeled samples. A sample can be mislabeled if there is label noise (i.e. negative margin), or if the perturbed sample is mislabeled (i.e. positive margin, but smaller than $\varepsilon$). Specifically, we define $\varepsilon_i$, the adversarial strength that is used to perturb the sample $i$ as:
\begin{equation}
    \varepsilon_i = \begin{cases}
        -\varepsilon & m_i \le -\varepsilon \\
        m_i & -\varepsilon < m_i < \varepsilon \\
        \varepsilon & m_i \ge \varepsilon \\
    \end{cases}
    \label{eq:eps_margin}
\end{equation}
where $\varepsilon$ is the default perturbation strength we want our model to be robust against, and $m_i$ is the model margin of the sample $i$. When $\varepsilon_i$ is a negative value, we use an anti-adversarial perturbation instead. That is, we apply a perturbation of value $|\varepsilon_i|$ to the sample to minimizes the loss instead of maximizing it.

\begin{figure*}[t]
    \centering
    \includegraphics[width=0.35\linewidth]{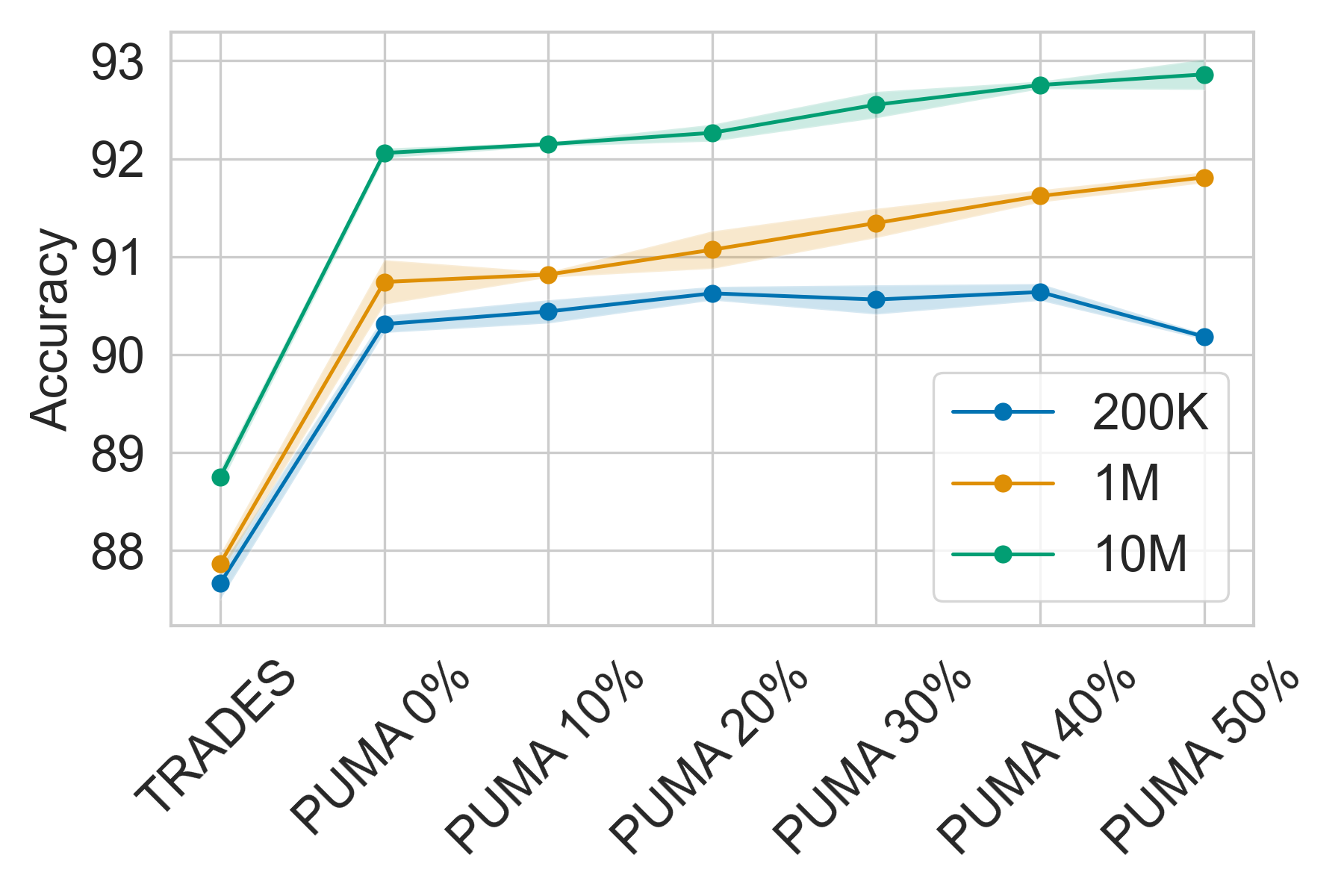}
    \includegraphics[width=0.35\linewidth]{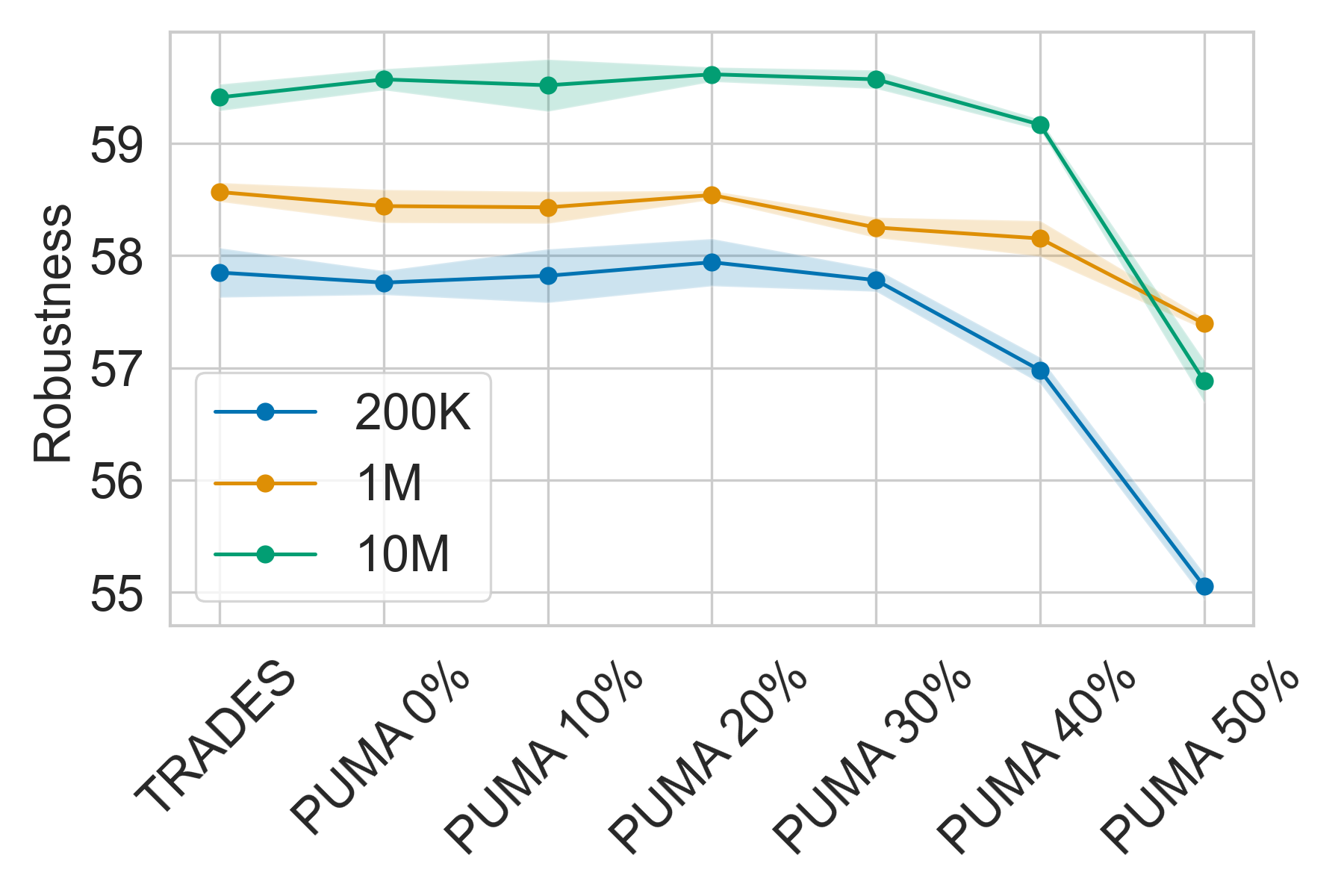}
    \caption{Performance of PUMA at different pruning ratios when training on CIFAR10 with either 200K, 1M or 10M EDM-generated examples. We use the ResNet-18 architecture in all cases.}
    \label{fig:data_size}
\end{figure*}

\begin{figure*}[t]
    \centering
    \includegraphics[width=0.35\linewidth]{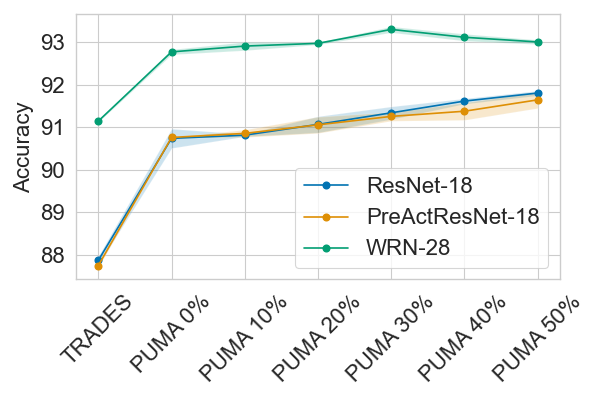}
    \includegraphics[width=0.35\linewidth]{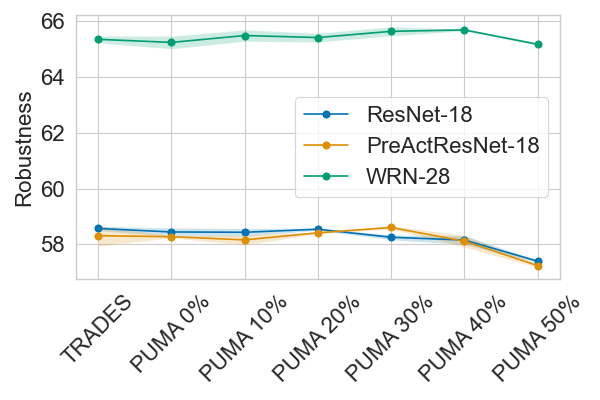}
    \caption{Performance of PUMA at different pruning ratios for the ResNet-18, PreActResnet-18 and WideResNet-28 architectures. We use the CIFAR10 dataset with 1M EDM-generated examples.}
    \label{fig:models}
\end{figure*}

\subsection{Performance}
\label{sec:performance}

In this Subsection, we provide empirical results that confirm the effectiveness of our design choices for different training sizes and architectures, and studies the model performance as we increase the pruning ratio. %We show that while our adaptive $\varepsilon$ is mostly responsible for the increment in accuracy, pruning the high-margin samples is still beneficial and reduces the data requirements.
In Figure \ref{fig:data_size}, we show that PUMA has significantly more accuracy than the SOTA model trained with TRADES~\cite{wang2023better}. While most of the increment in performance is thanks to our proposed per-sample $\varepsilon_i$, we show that pruning the highest margin data can increase the model accuracy, especially for larger quantities of data (1M and 10M samples) thanks to the larger $\alpha$, and may increase slightly the model robustness. This is consistent with our perceptron learning task findings, and contrasts with the general intuition that more data translates into better performance. However, we find a sharp decrease in robustness at higher pruning ratios. We believe this decrease is due to misalignment of our margin estimate with the ground truth, since it happens at the same pruning ratio for different quantities of additional data.

%\paragraph{Dataset}

In Figure \ref{fig:models}, we show that our previous results generalize to other architectures. We tested three architectures commonly used in image classification in the robustness literature. For a fair comparison, we used the same architecture to estimate the margin. We find that the ResNet-18 and PreActResNet-18 models behave very similarly, since they have a similar number of parameters. However, the much larger WideResNet-28 maximizes accuracy at a lower pruning ratio, because it has a significantly smaller $\alpha$ value.\looseness=-1

Finally, in Table~\ref{tab:puma_imagenet}, we show how our method performs in a more complex image classification task, for which we have images of higher resolution (i.e., 49 times the number of pixels than the CIFAR10 images). For the dataset, we propose to a variant of the ImageNet-21K dataset~\cite{ridnik2021imagenet21k} that has 100 classes. More details on how we constructed the dataset, how we modified the margin computation to be more scalable to the larger number of classes and training samples of this scenario, and how we trained the model with both PUMA and the state-of-the-art baseline~\cite{wang2023better}, can be found in Appendix~\ref{sec:puma_appendix/imagenet_details}. Despite the change in the dataset for this more complex task, we find similar results to what we obtained with CIFAR10. We find an even larger increase in accuracy from using PUMA over the SOTA baseline, and a larger increase when doing pruning. However, using PUMA in this dataset leads to a noticeable decrease in robustness, which is even more significant when we use relatively small pruning ratios. We believe this decrease in robustness is due to the much smaller margins that all the samples of this dataset have. In fact, more than 80\% of them have a margin value that is smaller than $\varepsilon = 8/255$, showing that properly making a model robust in this task is much more challenging. We think that larger models and more training time may be required in this scenario to maintain robustness when using PUMA and increasing the pruning ratio.

\begin{table}[t]
	\centering
	\begin{tabular}{ccc}
		\toprule
		& Accuracy & Robustness \\
		\midrule
		TRADES & $38.28$ & $\textbf{14.56}$ \\
		PUMA 0\% & $\textbf{43.69}$ & $13.05$ \\
		PUMA 10\% & $\textbf{44.66}$ & $7.22$ \\
		\bottomrule
	\end{tabular}
	\captionof{table}{Performance of PUMA with two different pruning ratios for our version of the ImageNet-21K dataset with 100 classes. We use the PreActResNet-18 architecture.}
	\label{tab:puma_imagenet}
\end{table}

\section{Model Analysis}
\label{sec:ablation}

% All the ablation study maybe moving epsilon buffer and  model used to estimate the margin subsections to the appendix.
We showed in Section~\ref{sec:puma} that PUMA can significantly increase the model accuracy without compromising the model robustness and while reducing data requirements. In this Section, we explore how the inner variables of PUMA can affect the final performance. Mainly, we discuss the effect of using a slightly smaller value of $\varepsilon_i$ to account for the margin estimation error, the effect of increasing the minimum and maximum values of $\varepsilon_i$, and we propose an online training method that avoids the main limitation of PUMA: having to precompute the margin of the training samples. 

Through this Section, we used an adversarially trained model on the original CIFAR10 dataset, without additional generated data, to compute the margin $m_i$ for PUMA. We analyze the effect of changing the architecture and the quantity of generated data used to train this model in Appendix~\ref{sec:model_margin}. Unless otherwise stated, we use 20\% for the pruning ratio.

\paragraph{Adjusting the $\varepsilon_i$ values}

One of the main principles of PUMA is to reduce the number of potentially mislabeled perturbed examples. However, because the model margin is a proxy of the true class boundary margin, using Eq.~(\ref{eq:eps_margin}) does not guarantee that no perturbed samples are mislabeled. We explore a simple solution: to further reduce $\varepsilon_i$ by a fixed quantity, or gap, to account for error in the margin estimation. That is, we use the following formula to compute $\varepsilon_i$:\looseness=-1
\begin{equation}
    \varepsilon_i = \begin{cases}
        -\varepsilon & m_i \le -\varepsilon + g \\
        m_i - g & -\varepsilon + g < m_i < \varepsilon + g\\
        \varepsilon & m_i \ge \varepsilon + g\\
    \end{cases}
    \label{eq:eps_margin_gap}
\end{equation}
where $g$ is the imposed gap between margin and adversarial strength.
We show in Table \ref{tab:gap} that imposing a gap results on reduced robustness and increased accuracy. We think that the reduction in the number of mislabeled perturbed examples is not significant enough to have a positive effect in robustness. Instead, because of the overall reduction in the $\varepsilon$ value, the model is trained on samples that are less adversarial, which reduces naturally the robustness and increases the accuracy.

\begin{table}[t]
    \centering
    \begin{tabular}{c|cc}
    \toprule
        & Accuracy & Robustness \\
        \hline
        No gap & $90.65_{\pm .09}$ & $\textbf{58.42}_{\pm .02}$ \\
        Gap = 1/255 & $\textbf{90.80}_{\pm .05}$ & $58.05_{\pm .07}$ \\
    \end{tabular}
    \caption{Performance results of using a gap (Eq.~\ref{eq:eps_margin_gap}) to make $\varepsilon_i$ smaller for the ResNet-18 model adversarially trained on CIFAR10 with 1M EDM-generated examples.}
    \label{tab:gap}
\end{table}

An additional change we can make to our method is increasing or decreasing the number of samples with adversarial strength determined by the margin. That is, we vary the $\varepsilon$ value in Eq.~(\ref{eq:eps_margin}). We observe in Table \ref{tab:epsilon} that we obtain a similar trade-off effect as in the previous experiment. Because most samples have positive margin, by increasing (or decreasing) the $\varepsilon$ value, we are effectively increasing (or decreasing) the adversarial strength, with its associated cost to accuracy or robustness.

\begin{table}[t]
    \centering
    \begin{tabular}{c|cc}
    \toprule
        & Accuracy & Robustness \\
        \hline
        $\varepsilon = 6/255$ & $\textbf{91.54}_{\pm .06}$ & $56.71_{\pm .15}$ \\
        $\varepsilon = 8/255$ & $90.65_{\pm .09}$ & $\textbf{58.42}_{\pm .02}$ \\
        $\varepsilon = 10/255$ & $90.04_{\pm .02}$ & $\textbf{58.55}_{\pm .32}$ \\
    \end{tabular}
    \caption{Performance results of using a larger (smaller) $\varepsilon$ in Eq.~(\ref{eq:eps_margin}) for the ResNet-18 model adversarially trained on CIFAR10 with 1M EDM-generated examples.}
    \label{tab:epsilon}
\end{table}

%\paragraph{Effect of pruning}

%Thus, our method PUMA effectively does two types of pruning: it removes the samples with the highest margin, but also reduces the attack strength on the samples with low margin. In Table \ref{tab:pruning_effect},

% \begin{table}[t]
%     \centering
%     \begin{tabular}{c|cc}
%     \toprule
%         & Accuracy & Robustness \\
%         \hline
%         PUMA & $\textbf{90.65}_{\pm .09}$ & $\textbf{58.42}_{\pm .02}$ \\
%         PUMA without pruning & $89.91_{\pm .14}$ & $\textbf{58.18}_{\pm .27}$ \\
%     \end{tabular}
%     \caption{Pruning is helpful in combination with tuning the adversarial strength.}
%     \label{tab:pruning_effect}
% \end{table}

% TODO: rewrite to be more clear (there is no previous point if the margin model is not here)
\paragraph{Online training} We explore if we can avoid the required precomputed margin of the samples. That is, we compute the margin during the PUMA algorithm, in an online manner.
To do this, we modify the adversarial attack. First, we obtain the perturbed sample $\vx'_i$. Second, we find if the images should be pruned or not. For that, we evaluate if the model prediction is correct for the same value of margin, $m_P$, that we used in the offline version to determine if the sample is removed or not. That is, we remove the sample if $p(\vx_i + \dfrac{m_i}{\varepsilon}(\vx'_i - \vx_i))) = y_i$. Third, we determine the adversarial strength using the margin value. Instead of computing the exact value, we solve the following optimization, which is constrained to the segment of relevant margin values to find $\varepsilon_i$ in Eq.~(\ref{eq:eps_margin}).\looseness=-1
\begin{equation}
     \max_{m_i} p(\vx_i + \dfrac{m_i}{\varepsilon}(\vx'_i - \vx_i))) = y_i \quad \forall m_i \in [-\varepsilon, \varepsilon]
\end{equation}
where $p(\vx) = \argmax(f(\vx))$. If there is no solution, we set the margin $m_i$ to $-\varepsilon$. Because of the constraint on $m_i$, we solve the maximization with binary search. We note that this $\varepsilon_i$ cannot be found incrementally like in IAAT~\cite{balaji2019instance} or CAT~\cite{cheng2020cat}, because the samples are not revisited multiple times, due to the large quantity of data available.\looseness=-1

%TODO: Write why online method fails
We see in Table~\ref{tab:online} the performance of the proposed online version compared to PUMA, the offline version. On the one hand, when the pruning ratio is 0\%, the online method performs well. The lower robustness and higher accuracy are probably due to the $\varepsilon_i$ values of difficult samples being relatively smaller at the beginning of the training. On the other hand, for easier samples, the $\varepsilon_i$ values are relatively large. Applying the same margin threshold as the offline method removes too many samples, and the model does not converge.\looseness=-1

While the proposed online method probably can have comparable performance to PUMA with lower computational cost, it is also more difficult to train because of the large variance of the margin at the beginning of the training and does not have a well-defined pruning ratio. We think a better online method may be achieved, by not pruning during the first training epochs.\looseness=-1

\begin{table}[t]
    \centering
    \begin{tabular}{c|cc}
    \toprule
        & Accuracy & Robustness \\
        \hline
        PUMA 0\% & $89.91_{\pm .14}$ & $\textbf{58.18}_{\pm .27}$ \\
        Online without pruning & $90.87_{\pm .03}$ & $57.49_{\pm .20}$ \\
        Online pruning $\varepsilon_i > 26/255$ & 14.87$_{\pm 6.99}$ & 12.8$_{\pm 4.06}$ \\
    \end{tabular}
    \caption{Performance of PUMA when using the online version of the method for the ResNet-18 model adversarially trained on CIFAR10 with 1M EDM-generated examples.}
    \label{tab:online}
\end{table}

% TODO: add margin correlation between margin models

\section{Conclusion}

In this work, we propose PUMA, a pruning strategy that uses a pretrained robust model, the margin model, to compute the margin of each training sample with DeepFool. Then, we remove the samples with the highest margin and compute the adaptive $\varepsilon_i$ values. Finally, we adversarially train a model using the adaptive $\varepsilon_i$ on the pruned training data. Not only we achieve our objective of reducing significantly the training size without reducing the resulting model robustness, but we are also able to improve its accuracy, surpassing SOTA robust models. 

We believe this work could help make the shift between data quantity to data quality in robust model training.
We think a promising direction would be to design generative models (e.g., diffusion models) that are conditioned on the margin. 
This way, given a particular budget, we can focus on generating images that have a lower overall margin, for better computation/performance tradeoffs.

\section*{Impact statement}

This paper presents work whose goal is to advance the field of Machine Learning. There are many potential societal consequences of our work, none which we feel must be specifically highlighted here.

\bibliography{references}
\bibliographystyle{icml2024}

\newpage
\appendix
\onecolumn

\begin{table}[t]
\centering
    \begin{minipage}[t]{0.48\linewidth}
    \centering
    \begin{tabular}{c|cc}
    \toprule
        & Accuracy & Robustness \\
        \hline
        $\varepsilon_i$ = $\varepsilon$ = 8/255 & $88.87_{\pm .10}$ & $57.40_{\pm .17}$ \\
        $\varepsilon_i$ = 1/255 if CD$_i$ = 1& $89.70_{\pm .08}$ & $57.87_{\pm .08}$ \\
        $\varepsilon_i \propto$ margin if CD$_i$ = 1 & $\textbf{90.66}_{\pm .01}$ & $\textbf{58.15}_{\pm .01}$ \\
        $\varepsilon_i \propto$ margin & $\textbf{90.67}_{\pm .14}$ & $\textbf{58.15}_{\pm .08}$ \\
    \end{tabular}
    \captionof{table}{Performance with different ways of changing the adversarial attack strength for the ResNet-18 model adversarially trained on CIFAR10 with 1M EDM-generated examples. In all cases we use the PE strategy, pruning the samples with CD $<$ 0.02. The most effective approach is to select $\varepsilon_i$ to be proportional to the margin as in Eq.~(\ref{eq:eps_margin}), regardless of the sample CD.}
    \label{tab:margin_difficult}
\end{minipage}
\hfill
    \begin{minipage}[t]{0.48\linewidth}
    \begin{tabular}{c|cc}
    \toprule
        & Accuracy & Robustness \\
        \hline
        Random pruning 22\% & $87.64_{\pm .04}$ & $58.27_{\pm .05}$ \\
        PUMA 22\% - CD pruning & $\textbf{90.67}_{\pm .14}$ & $58.15_{\pm .08}$ \\
        PUMA 22\% & $\textbf{90.65}_{\pm .09}$ & $\textbf{58.42}_{\pm .02}$ \\
    \end{tabular}
    \captionof{table}{Effects of changing the pruning metric for the ResNet-18 model adversarially trained on CIFAR10 with 1M EDM-generated examples. We prune 22\% of samples to be able to compare with CD-based pruning. highest margin results in similar robustness as random pruning, but with 3 points more of accuracy.}
    \label{tab:pruning_strategies}
\end{minipage}
\end{table}

\section{CD instead of model margin}
\label{sec:cd_margin_comparison}

In Table \ref{tab:margin_difficult}, we show the potential of using the margin to adjust the adversarial strength. With respect to only pruning the samples with CD $<$ 0.02 (i.e., 22\% of the data), we can significantly improve the accuracy and robustness by using weaker attacks. Our results, corroborate our previous intuitions: the most difficult samples tend to be mislabelled when perturbed, and require weaker attacks. In fact, just weakening the training attack on the most difficult samples (CD = 1) is able to improve both the accuracy and robustness of the model. However, we obtain our best results by using an attack strength that is proportional to the margin, as in Eq.~\ref{eq:eps_margin}. Moreover, we see that the latter approach has similar results when applied on the most difficult samples or any sample in general.

Seeing the relative success of using the model margin to adjust the training attack $\varepsilon$, we ask ourselves if the CD metric is really necessary for our method. Unlike previous works, our method relies on the margin, so it would be preferable if we could use this metric for both adjusting the adversarial strength as well as the pruning. In Table \ref{tab:pruning_strategies} we show the difference in results when using the CD metric to prune the easy samples compared to using the margin itself. In both cases, we have used our approach to reduce the attack strength of the samples with lowest margin, using Eq.~(\ref{eq:eps_margin}). We can see that we can not only remove the CD metric from our method, but it is even preferable to do so. Moreover, compared to fixing the epsilon value and using random pruning, our PUMA method has similar robustness but considerably higher accuracy.

\begin{table}[t]
    \centering
    \begin{tabular}{l|cc}
    \toprule
         Margin model (MM) & Accuracy & Robustness \\
        \hline
        ResNet-18 TRADES on CIFAR10 & $90.65_{\pm .09}$ & $\textbf{58.42}_{\pm .02}$ \\
        ResNet-18 TRADES on CIFAR10+1M (*) & $\textbf{91.37}_{\pm .21}$ & $\textbf{58.53}_{\pm .36}$ \\
        ResNet-18 PUMA using (*) as margin model & $\textbf{91.20}_{\pm .06}$ & $58.24_{\pm .01}$ \\
        WRN-28 TRADES on CIFAR10+50M (NC) & 89.70$_{\pm .11}$	& 56.34$_{\pm .02}$ \\
        WRN-28 TRADES on CIFAR10+1M & 90.65$_{\pm .02}$ & 57.37$_{\pm .12}$ \\
        WRN-70 TRADES on CIFAR10+50M (NC) & 88.83$_{\pm .13}$	& 54.80$_{\pm .02}$ \\
    \end{tabular}
    \caption{Performance of a ResNet-18 trained on CIFAR10 with 1M generated data samples with PUMA when we change the architecture, training method, and training data of the model used to estimate the margin. ''NC": the 1M samples are not contained in the 50M.\looseness=-1}
    \label{tab:margin_model}
\end{table}

\section{Changing the model used to compute the margin} 
\label{sec:model_margin}
% TODO: change based on new results

% TODO: main message? not super clear
Our method heavily relies on the margin estimate to decide the optimal attack parameters for each of the samples. In Section~\ref{sec:ablation}, we have used a robust ResNet-18, trained only on the original CIFAR10 data to estimate the margin, but our best results use the general settings of Section~\ref{sec:puma}, in which we have used a robust ResNet-18 trained on the CIFAR10 data with 1M EDM-generated examples. In this Section, we explore how the performance of the PUMA-trained model changes depending on the architecture, training method and training data used for the pretrained model that computes the margin estimate with DeepFool. We will refer to the latter model as margin model (MM).

In Table~\ref{tab:margin_model}, we show the performance when using different margin models to compute the margin estimates for a ResNet-18 trained with PUMA on the CIFAR10 dataset with 1M generated data examples. First, we find that the best result is achieved when the architecture and training data are the same as the model we train PUMA with. When we change the margin model architecture to the better WRN-28, the performance decreases. Similarly, the performance decreases when the dataset changes, even if the performance of the margin model increases by adding more data (50M examples in our case). Thus, we observe that a mismatch in data or architecture between the two models results in decreased final performance.

Finally, we have also explored using our method iteratively, as shown in the third row of Table~\ref{tab:margin_model}. That is, we compute our estimates with the model margin that achieves the best result, train a model with PUMA on those estimates, and then recompute the estimates with the trained model. Using those estimates to train a model with PUMA yields slightly worse results, thus removing the potential of refining the margin estimates.

\section{Experimental details of the ImageNet-21K results}
\label{sec:puma_appendix/imagenet_details}

In this Section, we will provide more details on how we constructed our variant of the ImageNet-21K dataset~\cite{ridnik2021imagenet21k} that we use in Subsection~\ref{sec:performance}, details on how we modified the margin computation to be more scalable to the larger number of classes and training samples of this scenario, and details on how we trained the model with both PUMA and the state-of-the-art baseline~\cite{wang2023better}. 

For the dataset, we propose to use a variation of the ImageNet-21K dataset, in which we use only 100 classes to maximize the number of images per class and reduce class imbalance. Unlike the much more popular ImageNet-1K~\cite{krizhevsky2012imagenet}, which only has about 1000 images per class, the ImageNet-21K dataset has ten times the quantity of data, and detailed class hierarchy data. This hierarchy allows us to group up classes of objects that belong to bigger categories (e.g., breeds of dogs), and effectively increase the number of images per class. Increasing the number of data samples per class is necessary because the potential improvement of our method is dependent on the quantity of data, as we showed in Subsection~\ref{sec:performance}. To maximize the number of images per class, we first use the top superclasses (e.g., animal, food, sport,...), then we iteratively divide the most popular one into its subcategories until we have 100 classes with similar number of images (at most five times the difference from the most popular to the least popular) and finally, remove the rest. This results in a dataset with approximately 5.3 million images, giving us around 50 times the number of samples per class than the ImageNet-1K dataset.\looseness=-1

\begin{figure}[t]
	\centering
	\begin{subfigure}{0.4\linewidth}
		\centering
		\includegraphics[width=\linewidth]{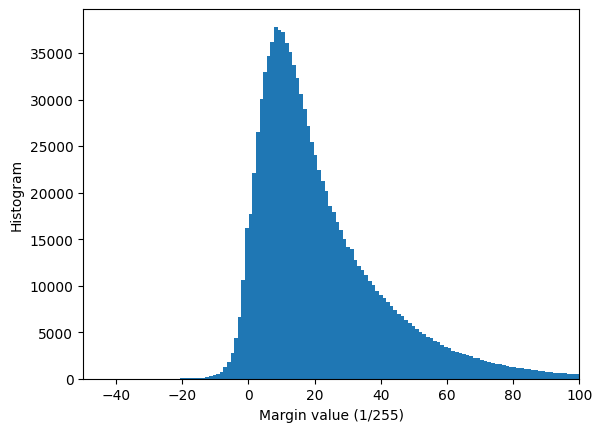}
		\caption{Using DeepFool}
	\end{subfigure} 
	\begin{subfigure}{0.4\linewidth}
		\centering
		\includegraphics[width=\linewidth]{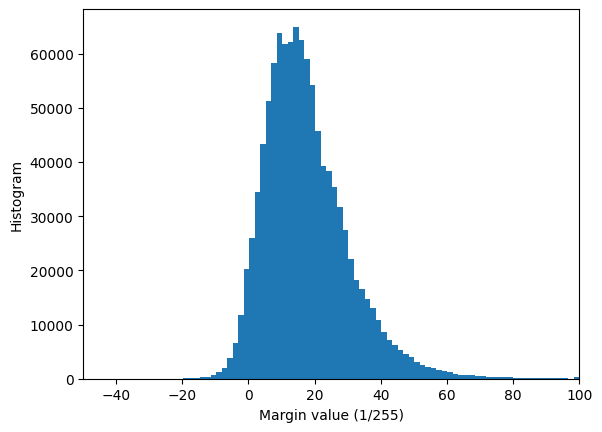}
		\caption{Using Algorithm~\ref{alg:faster_margin}}
	\end{subfigure}
	\caption{Histogram of the margin values of the CIFAR10 training set with 1M additional EDM-generated samples, when computed with DeepFool and our proposed faster margin algorithm.}
	\label{fig:margin_distribution}
\end{figure}

For the margin computation, we find that using DeepFool in ImageNet-21K is no longer computationally efficient. The main reason being that DeepFool needs to compute the trace of the Jacobian to craft minimal targeted adversarial attacks, making its cost scale linearly with the number of classes. Instead, we propose Algorithm~\ref{alg:faster_margin}, which is partially inspired by the online training method we proposed in Section~\ref{sec:ablation}. The algorithm has two main steps. First, we check if the model classifies correctly the natural sample (i.e., not adversarially perturbed). Then, we use BIM~\cite{kurakin2017adversarial} (i.e., PGD without random initialization) to either craft an adversarial perturbation if the sample was classified correctly, or an anti-adversarial perturbation otherwise. However, we stop it early as soon as the model misclassifies or correctly classifies, respectively. Measuring the norm of this perturbation gives us a good estimate of the margin, where in the case of anti-adversarial perturbation it would be negative. Finally, we use binary search on the segment from this adversarial example to the natural sample, to refine the margin value. In Figure~\ref{fig:margin_distribution}, we show that, compared to DeepFool, this method gives relatively similar margin distributions in the CIFAR10 data, and thus, it is a promising option to generate the margins in a much more computationally efficient manner.

\begin{algorithm}[t]
	\caption{Faster margin algorithm for ImageNet datasets
		\label{alg:faster_margin}}
	\begin{algorithmic}[1]
		\STATE {\bfseries Input:} Classifier $f$, Image $\vx$, label $y$ and the precision parameters $\alpha$, $i_{max}$, $j_{max}$
		\STATE {\bfseries Output:} Margin value $m$
  \STATE 
		\STATE Initialize: $\vx' \gets \vx, p_0 \gets f(\vx), i \gets 1, k \gets 1$
		\IF{$p_0 \neq y$}
		\STATE{$k \gets -1$}
		\ENDIF
		\WHILE{$(f(\vx') = p_0) \text{ and } (i \leq i_{max})$}
		\STATE{$\vx' \gets \vx' + k \cdot \alpha \sign (\nabla_{\vx} \text{CE}(f(\vx'), y))$} \COMMENT{CE: cross-entropy loss}
		\STATE{$i \gets i + 1$}
		\ENDWHILE
		\STATE{$m \gets k \lVert \vx' - \vx \rVert_{\infty}$}
		\STATE{$m_{down} \gets \min(0,m), m_{up} \gets \max(0,m)$}
		\FOR{$j \gets 1 \textrm{ to } j_{max}$}
		\STATE{$m \gets (m_{down} + m_{up})/2$}
		\STATE{$p \gets f(\vx + (\vx' - \vx) \cdot m)$}
		\IF{$p = y$}
		\STATE{$m_{down} \gets m$}
		\ELSE
		\STATE{$m_{up} \gets m$}
		\ENDIF
		\ENDFOR
		\STATE{$m \gets (m_{down} + m_{up})/2$}
	\end{algorithmic}
\end{algorithm}

For the ImageNet-21K experiments with the SOTA baseline and PUMA methods we use the PreActResNet-18 architecture (PRN-18) for the model. For the PUMA algorithm, we use the PRN-18 trained with the SOTA baseline and compute the margin with the aforementioned algorithm. In all the cases, we will use the same training hyperparameters. We use stochastic gradient descent (SGD) for 107 epochs, where each epoch is composed of 50000 training samples, (i.e., one pass through each training sample). We use the OneCycle learning rate with two phases, cosine annealing and 0.2 as the maximum learning rate, which is updated every batch. For the batch size we use 64 samples in both training and testing due to memory constraints. All adversarial examples are crafted with L$_{\infty}$ perturbations of maximum size $\varepsilon = 8/255$. We craft the training perturbations with PGD-10 with step size 2/255, and the test robustness with C\&W attacks~\cite{carlini2017towards}.\looseness=-1

\end{document}